%% file: main.tex
\documentclass[10pt,twocolumn,letterpaper]{article}

\usepackage{iccv}      


\definecolor{iccvblue}{rgb}{0.21,0.49,0.74}
\usepackage[pagebackref,breaklinks,colorlinks,allcolors=iccvblue]{hyperref}
\usepackage{tcolorbox}
\tcbuselibrary{breakable} 
\tcbuselibrary{skins} 
\usepackage{multirow,multicol}

\tcbset{
    userstyle/.style={
        enhanced,
        colback=white,
        colframe=black,
        colbacktitle=gray!20,
        coltitle=black,
        rounded corners,
        sharp corners=north,
        boxrule=0.5pt,
        drop shadow=black!50!white,
        attach boxed title to top left={
            xshift=-2mm,
            yshift=-2mm
        },
        boxed title style={
            rounded corners,
            size=small,
            colback=gray!20
        }
    },
    replystyleg/.style={
        enhanced,
        colback=green!15,
        colframe=black,
        colbacktitle=green!30,
        coltitle=black,
        boxrule=0.5pt,
        drop shadow=black!50!white,
        rounded corners,
        sharp corners=north,
        attach boxed title to top right={
            xshift=-2mm,
            yshift=-2mm
        },
        boxed title style={
            rounded corners,
            size=small,
            colback=green!40
        }
    },
    replystyler/.style={
        enhanced,
        colback=red!15,
        colframe=black,
        colbacktitle=red!40,
        coltitle=black,
        boxrule=0.5pt,
        drop shadow=black!50!white,
        rounded corners,
        sharp corners=north,
        attach boxed title to top right={
            xshift=-2mm,
            yshift=-2mm
        },
        boxed title style={
            rounded corners,
            size=small,
            colback=red!40
        }
    }
}

\newtcolorbox{userquery}[1][]{
    userstyle,
    title=Prompt,
    #1
}

\newtcolorbox{llmreply-g}[1][]{
    replystyleg,
    title=Response,
    #1
}

\newtcolorbox{llmreply-r}[1][]{
    replystyler,
    title=Response,
    #1
}
\def\paperID{12030} 
\def\confName{ICCV}
\def\confYear{2025}

\title{Mitigating Object Hallucinations in MLLMs via Multi-Frequency Perturbations}

\author{Shuo Li\thanks{{ }Equal contributions.}$^{  \ ,1}$, Jiajun Sun$^{*,1}$, Guodong Zheng,
Xiaoran Fan$^{1}$, Yujiong Shen$^{1}$,\\
Yi Lu$^{1}$, Zhiheng Xi$^{1}$, Yuming Yang$^{1}$,
Wenming Tan$^{2}$, Tao Ji$^{\dag,  1 }$, Tao Gui\thanks{{ }Corresponding author.}$^{ \ ,1}$, Qi Zhang$^{1}$, Xuanjing Huang$^{1}$\\
Fudan University$^{1}$  \  \ Hikvision Research Institute$^{2}$\\
\texttt{lis23@m.fudan.edu.cn, \{taoji, tgui\}@fudan.edu.cn} 
}

\begin{document}
\maketitle

\input{sec/0_abstract}

\input{sec/1_Introduction}

\input{sec/2_RelatedWork}

\input{sec/3_Method}

\input{sec/4_Experiment}

\input{sec/5_Conclusion}

{
    \small
    \bibliographystyle{ieeenat_fullname}
    \bibliography{main}
}

\appendix
\input{sec/appendix}

\end{document}

%% file: sec/0_abstract.tex
\begin{abstract}
Recently, multimodal large language models (MLLMs) have demonstrated remarkable performance in visual-language tasks. However, the authenticity of the responses generated by MLLMs is often compromised by object hallucinations. We identify that a key cause of these hallucinations is the model’s over-susceptibility to specific image frequency features in detecting objects. In this paper, we introduce Multi-Frequency Perturbations (MFP), a simple, cost-effective, and pluggable method that leverages both low-frequency and high-frequency features of images to perturb visual feature representations and explicitly suppress redundant frequency-domain features during inference, thereby mitigating hallucinations. Experimental results demonstrate that our method significantly mitigates object hallucinations across various model architectures. Furthermore, as a training-time method, MFP can be combined with inference-time methods to achieve state-of-the-art performance on the CHAIR benchmark.~\footnote{Our code will be made publicly available.}

\end{abstract}

%% file: sec/1_Introduction.tex
\begin{figure*}[ht!]
  \centering
   \includegraphics[width=0.9\linewidth]{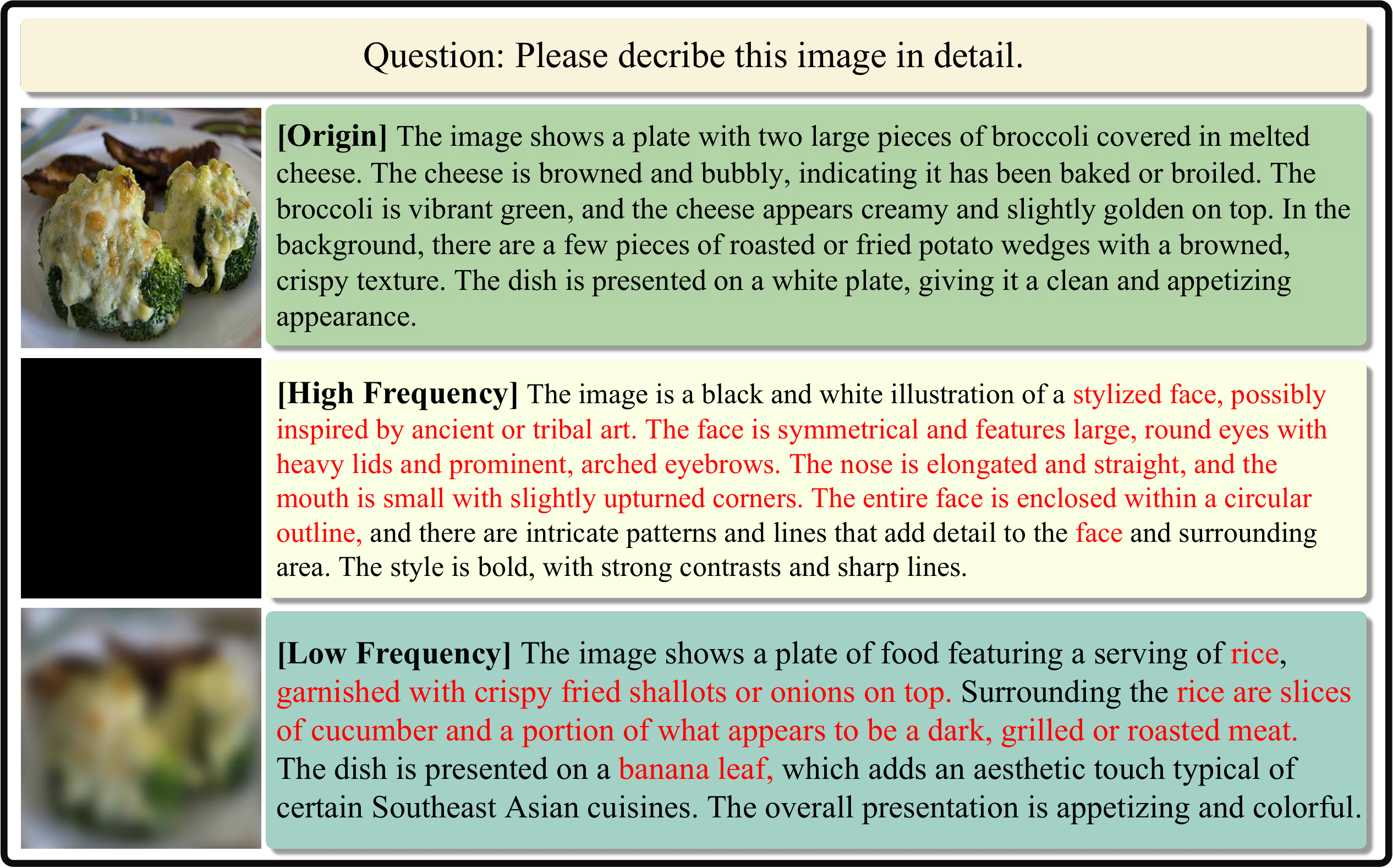}

   \caption{An example of GPT-4o. Unlike humans, the model is over-susceptible to limited high- and low-frequency image features to detect objects, leading to incorrect object detection and erroneous image caption.}
   \label{fig:into}
\end{figure*}

\begin{figure}
    \centering
    \begin{subfigure}{0.95\linewidth}
        \centering
        \includegraphics[width=\linewidth]{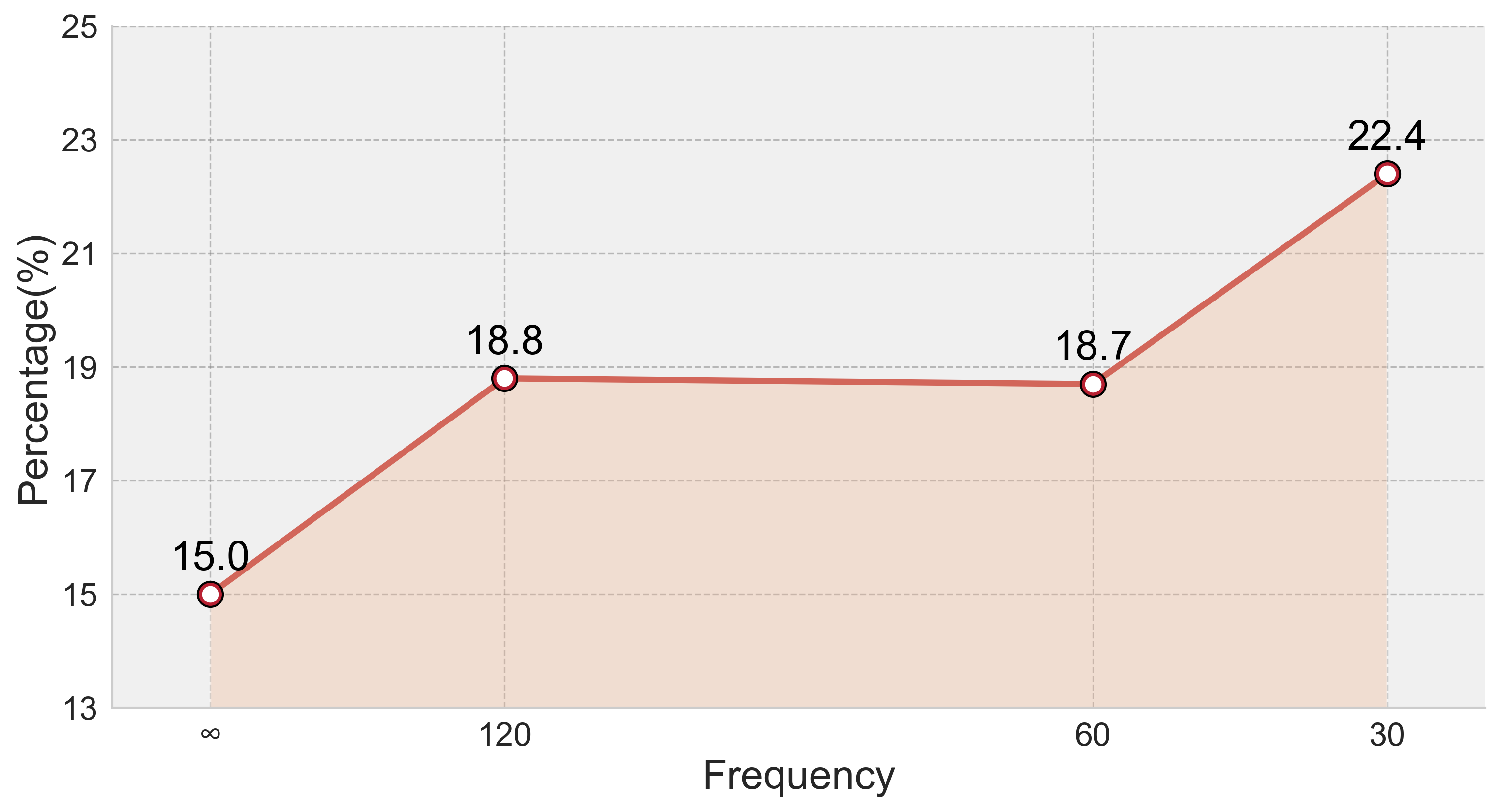}
        \caption{Low-frequency}
        \label{fig:sub1}
    \end{subfigure}
    
    \begin{subfigure}{0.95\linewidth}
        \centering
        \includegraphics[width=\linewidth]{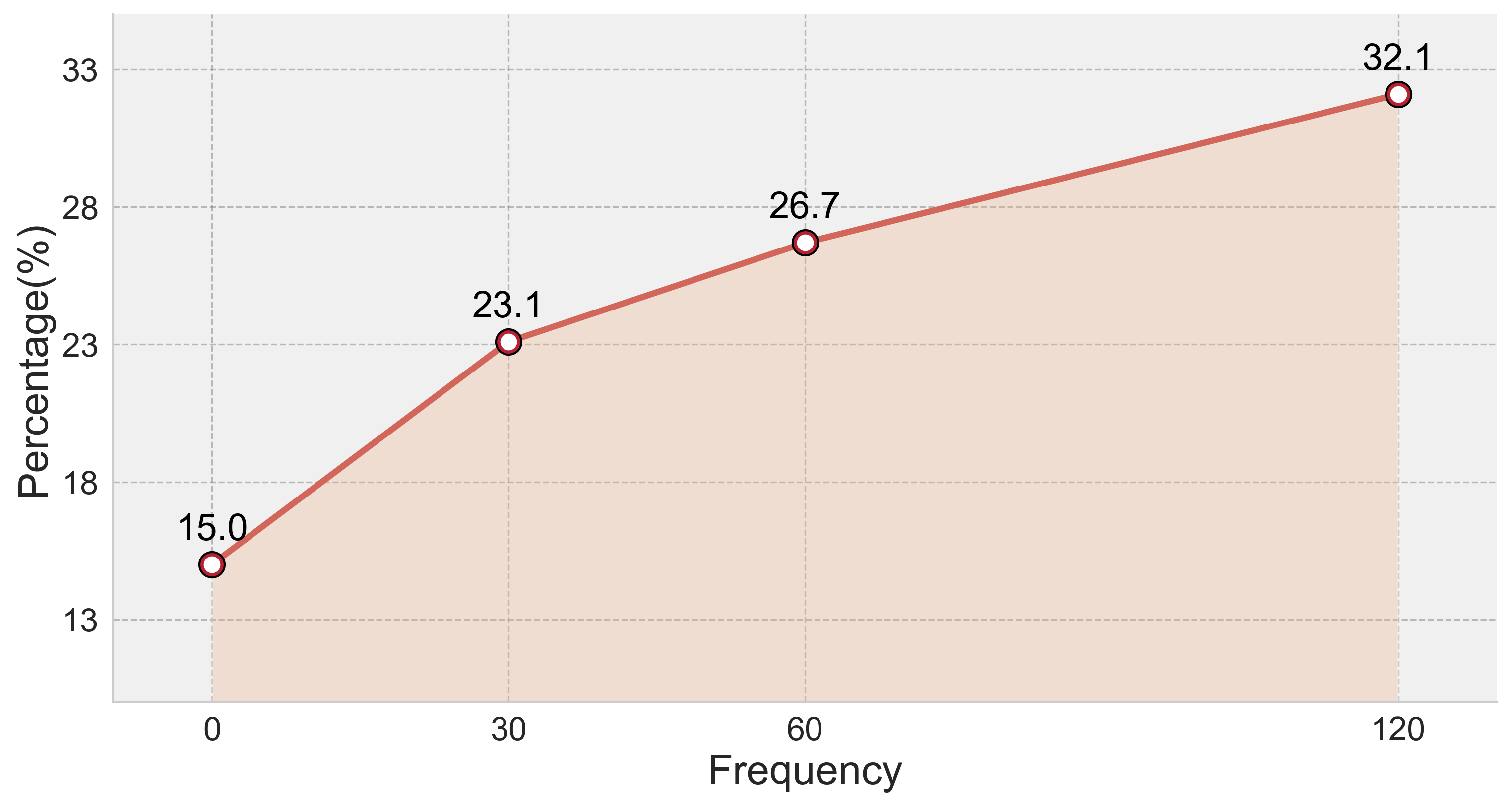}
        \caption{High-frequency}
        \label{fig:sub2}
    \end{subfigure}
    
    \caption{Instance-level hallucination rate when using only low or high frequency features. The x-axis represents the cutoff frequency. Features with frequencies higher than the cutoff are retained as high-frequency features, while those below the cutoff are selected as low-frequency features.}
    \label{fig:intro}
\end{figure}
\section{Introduction}
Large language models (LLMs), exemplified by ChatGPT~\cite{ChatGPT}, demonstrate remarkable performance across diverse text-based tasks. By integrating visual encoders like CLIP~\cite{radford2021learningtransferablevisualmodels}, multimodal large language models (MLLMs) extend these capabilities to visual domains. These models excel in a range of vision-language tasks, including image caption~\cite{wang2020overview}, visual question answering~\cite{antol2015vqa}, and visual dialogue~\cite{das2017visual}.

However, MLLMs always face the challenge of object hallucinations~\cite{rohrbach2018object,li2023evaluating}, where the model's outputs fail to accurately correspond to the objects in the real image. This issue undermines model performance and diminishes its credibility.

To tackle this issue, previous studies propose various methods~\cite{bai2024hallucination,lan2024survey}, which can be broadly categorized into training-time and inference-time methods. However, these approaches overlook a critical weakness in MLLMs. Specifically, the MLLMs tend to be over-susceptible to  low- or high-frequency features in the image to detect objects, often neglecting the actual features of the real image. For instance, as illustrated in~\cref{fig:into}, MLLMs can still identify objects in images, even when blurred images containing certain frequencies are used. This characteristic may contribute to the occurrence of object hallucination. This phenomenon has also been widely observed in studies on image frequency-domain attacks targeting vision models~\cite{wang2021backdoor,long2022frequency,feng2022fiba} in other models.

Based on these insights, we introduce Multi-Frequency Perturbations (MFP), a simple, cost-effective, and pluggable method that leverages both low-frequency and high-frequency features of images to perturb visual feature representations and explicitly suppress redundant frequency-domain features during inference. MFP partitions an image into high-frequency and low-frequency components, extracts their respective features, and performs a fine-grained fusion of these features with the original image features at the visual token level. This process results in more robust image features. The module is composed of two main components: (1) Multi-Frequency Feature Extraction. In this step, Gaussian high-pass and low-pass filters~\cite{young1995recursive} are applied to the original image to obtain its raw high-frequency and low-frequency features. These raw features are then fed into the visual encoder, producing the corresponding high-frequency and low-frequency features, represented as visual token sequences. (2) Fine-Grained Frequency Feature Fusion. We use the original visual token sequence derived from the image encoder along with the high-frequency and low-frequency token sequences. These are fused at the token level using a cross-attention mechanism to generate the final perturbed visual token sequences. During inference, we apply a decay to both high-frequency and low-frequency feature perturbations, thereby reducing redundant high-frequency and low-frequency features. Experimental results demonstrate that our method achieves remarkable effectiveness in MLLM object hallucination benchmarks, regardless of visual encoders, LLM backbones, resolutions, or size. Additionally, our training-time approach can be combined with inference-time methods to achieve better performance and even achieve SOTA results on CHAIR~\cite{rohrbach2018object}.

In this paper, our main contributions are: 
\begin{itemize}[leftmargin=*]
    \item we are the first to address the cause of object hallucinations from the frequency domain perspective and introduce MFP, a simple, efficient, and pluggable method that effectively mitigates object hallucination in MLLMs;
    \item we demonstrate the effectiveness of our proposed method across models of various architectures, highlighting its strong generalization capabilities;
    \item we demonstrate that our proposed training-time method can be combined with inference-time method to reach better results, even achieving state-of-the-art performance on the CHAIR benchmark.
\end{itemize}

%% file: sec/2_RelatedWork.tex
\section{Related Work}

\paragraph{Multimodal Large Language Models} MLLMs, represented by GPT-4~\cite{openai2024gpt4technicalreport}, have demonstrated remarkable capabilities and are rapidly becoming a key area of research in Deep Learning. By combining visual and language models, they enable cross-modal understanding and reasoning. Models like CLIP~\cite{radford2021learningtransferablevisualmodels} have bridged the gap between language models and visual tasks, showcasing the potential of cross-modal applications. The BLIP~\cite{li2022blip,blip2,dai2023instructblipgeneralpurposevisionlanguagemodels} series has advanced to support tasks such as visual question answering, while LLaVA~\cite{liu2024visual,liu2024llavanext} employs a simple linear projection layer and a two-stage training method to enhance image-text spatial alignment and overall model performance. Additionally, MouSi~\cite{fan2024mousi} and Cambrian-1~\cite{tong2024cambrian1fullyopenvisioncentric} utilize the strengths of diverse visual encoders to enrich multimodal understanding. Recently, the InternLM-XComposer~\cite{internlmxcomposer,internlmxcomposer2} and InternVL~\cite{chen2023internvl,chen2024far} families of models demonstrate leading performance, typically following an architecture similar to LLaVA.

\paragraph{Object Hallucinations in MLLMs} Object hallucinations in MLLMs occur when the model generates outputs related to objects that do not correspond to actual objects in the input image, leading to false or inaccurate visual associations. This misalignment often arises from insufficient cross-modal fusion~\cite{tong2024eyeswideshutexploring}, over-reliance on dataset biases~\cite{Yu_2024_CVPR}, and the model’s tendency to generate text-driven predictions rather than faithfully grounding outputs in visual inputs~\cite{liu2024paying}. Existing methods to address this issue can be broadly classified into training-time and inference-time methods. In training-time methods,~\cite{chen2023mitigating,jiang2024hallucination,yue2024less} focus on auxiliary supervision, while~\cite{zhao2023beyond,zhou2024aligning,sun2023aligning,ben2023mitigating,yu2024rlhf} using reinforcement learning to mitigate hallucinations. For inference-time methods,~\cite{leng2024mitigating,zhu2024ibd,zhao2024mitigating,huang2024opera,han2024skip} employ generative interventions to reduce object hallucinations, while~\cite{yin2024woodpecker,lee2023volcano,zhou2023analyzing} apply post-correction strategies. However, as we know, all these methods neglect the perspective of the frequency domain.

\paragraph{Visual Feature in Frequency Domain} 
In image representation learning~\cite{xu2020learning}, frequency domain features are extracted using methods such as fourier transform~\cite{bracewell1989fourier} and wavelet transform~\cite{zhang2019wavelet}. While RGB features represent the magnitude of image pixel values, frequency domain features capture the intensity of pixel value changes. High-frequency information highlights object edges, whereas low-frequency information defines the general outline of objects. Frequency domain features have been widely applied in areas such as camouflage object detection~\cite{cong2023frequency,lin2023frequency,xie2023frequency} and super-resolution restoration~\cite{guan2024frequency,behjati2022frequency}. However, to the best of our knowledge, there is still a lack of research~\cite{liu2024hiprompt} exploring the role of the visual feature in the frequency domain within MLLMs. Even no prior work explores their role in object hallucinations in MLLMs.

%% file: sec/3_Method.tex
\begin{figure*}[ht!]
  \centering
   \includegraphics[width=1\linewidth]{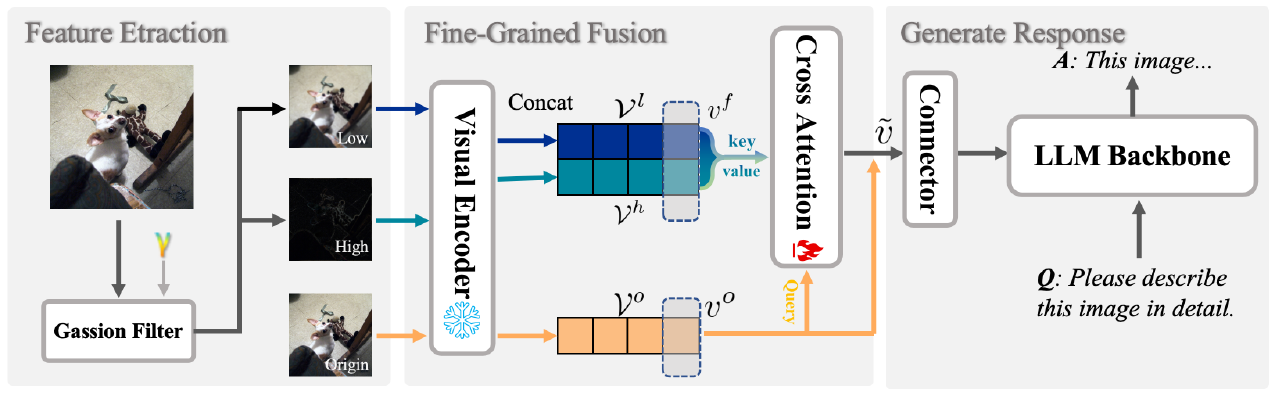}

   \caption{The model architecture of our proposed method. Where $\gamma$ is only employed at inference time.}
   \label{fig:main}
\end{figure*}
\section{Method}
In this section, we first demonstrate how MLLMs are overly susceptible to frequency domain features for object detection, leading to object hallucination. Next, we introduce our proposed method that applies multi-frequency perturbations. Finally, we introduce our setup in training and inference time.
\subsection{Over-Susceptibility to Frequency in MLLM}
Unlike most previous works that focus on the model's decoding or attention layers, we investigate which image features the model relies on for object detection. We seek to answer the question: \textit{Does the model establish the correct connections between image features and objects?}

As shown in~\cref{fig:main}, we apply filters to retain only the high- or low-frequency features of an image, causing significant distortion that makes the objects barely distinguishable. We then prompt the MLLMs with the instruction: ``Please describe this image in detail.'' However, we observe a common phenomenon: almost all of the MLLMs, including GPT-4o~\cite{hurst2024gpt}—one of the most advanced MLLMs—fail to recognize images as meaningless or devoid of objects. Instead, they often generate incorrect captions and hallucinate objects.

Next, we use quantitative methods to validate the existence of this phenomenon. As shown in~\cref{fig:intro}, we randomly selected 200 images from the MSCOCO dataset~\cite{lin2014microsoft}, retaining only a portion of the high- or low-frequency features. We then prompted the MLLMs to describe these images and calculated the proportion of instances containing object hallucinations (Recognizing images as meaningless or devoid of objects will not be classified as hallucination). We observed a significant increase in the proportion of hallucinatory sentences as less high-frequency feature was preserved (with the cutoff frequency ranging from 0 to 120), rising from 15.0 to 32.1. A similar trend was seen with low-frequency features (cutoff frequency from infinity to 30), where the proportion increased from 15.0 to 24.4.

Based on these results, we observe that MLLMs tend to be over-susceptible to low-frequency or high-frequency features in the image for detecting objects. However, this tendency makes MLLMs vulnerable and overly sensitive to interference from redundant low- or high-frequency  features, which can lead to hallucinations in the models.
\subsection{Multi-Frequency Perturbutaions}

\paragraph{Model Architecture}
Our model architecture, similar to most MLLMs, as shown in \cref{fig:main}, 
consists of three main components: a CLIP-like visual encoder, connectors, and a large Language Model (LLM). During training, the Gaussian filter is used to extract both high- and low-frequency features from the image. These features are then fine-grained and fused with the original image features. The resulting fused features as visual tokens are aligned with the text tokens via the connector. The concatenated tokens are then fed into the LLM backbone to generate the response.
\paragraph{Multi-Frequency Feature Extraction}
We use a Gaussian filter to extract the high-frequency and low-frequency features from the image. Specifically, let $\mathcal{I}^o(u,v) \in \mathbb{R}^{h\times w \times 3}$ denote the original image, an RGB three-channel image of height $u \in [0,h]$ and width $v \in [0,w]$. 
First, we apply the Fourier transform to each channel separately:
\begin{equation}
  \mathcal{F}_c(u,v) = FFT(\mathcal{I}_c(u,v)), \ c \in \{R,G,B\},
  \label{eq:important}
\end{equation}
then we define the Gaussian low-frequency filter $\mathcal{H}^l_c(u,v)$ and high-frequency filter $\mathcal{H}^h_c(u,v)$ as follows:
\begin{equation}
    \begin{cases}
  \mathcal{H}^l_c(u,v) = \exp\left(-\frac{\mathcal{D}^2(u,v)}{2\mathcal{D}_0^2}\right) \\
  \mathcal{H}^h_c(u,v) = 1 -\exp\left(-\frac{\mathcal{D}^2(u,v)}{2\mathcal{D}_0^2}\right),
    \end{cases}
\end{equation}
where $\mathcal{D}^2(u,v)$ represents the distance from the pixel point $(u,v)$ to the center of the frequency rectangle, and $\mathcal{D}_0$ denotes the cutoff frequency.These filters are then applied to filter the low and high frequency of the original image:
\begin{equation}
\label{eq:get_freq}
    \begin{cases}
  \mathcal{F}^l_c(u,v) = \mathcal{F}_c(u,v) \cdot \mathcal{H}^l_c(u,v) \\
  \mathcal{F}^h_c(u,v) = \mathcal{F}_c(u,v) \cdot \mathcal{H}^h_c(u,v).
    \end{cases}
\end{equation}
\noindent Finally, we apply the inverse Fourier transform to these frequencies:
\begin{equation}
    \begin{cases}
  \mathcal{I}^l_c(u,v) =FFT^{-1}( \mathcal{F}^l_c(u,v)) \\
  \mathcal{I}^h_c(u,v) = FFT^{-1}(\mathcal{F}^h_c(u,v)),
    \end{cases}
\end{equation}
and combine them into the RGB channels to obtain the low-frequency features $\mathcal{I}^l(u,v)$ and high-frequency features $\mathcal{I}^h(u,v)$ of the image.
\paragraph{Fine-Grained Frequency Feature Fusion}
After obtaining the low-frequency features $\mathcal{I}^l(u,v)$ and high-frequency features $\mathcal{I}^h(u,v)$ of the image, we fuse them to the origin image features using a cost-effective method. We then encode these features with the visual encoder $VG$(\eg, CLIP) to obtain a sequence of visual tokens $V \in \mathbb{R}^{L \times dim}$:
\begin{equation}
\mathcal{V}=\mathrm{VG}(\mathcal{I}),
\end{equation}
where $L$ depends on the settings of the visual encoder, and $dim$ refers to the dimension of the visual encoder's hidden layer.

For each token $v$ in the visual token sequence we obtained, we use cross-attention to fuse the high-frequency token $v^h$ and low-frequency visual token $v^l$ into the visual token $v^o$ of the original image at the same position in the sequence. This process can be expressed as follows:
\begin{align}
&v^o \in \mathcal{V}^o,v^l \in \mathcal{V}^l,v^h \in \mathcal{V}^h,\nonumber\\
&v^f = v^l \oplus v^h,\nonumber\\
&\tilde{v} = \mathrm{softmax}\left(\frac{v^oW^q(v^fW^k)^\top}{\sqrt{d_{k} } }\right)v^fW^v+v^o,
\label{eq:train_main}
\end{align}
where $\oplus$ represents the concatenation operation, $v_f \in \mathbb{R}^{2\times dim}$ is the multi-frequency token obtained by concatenating $v^l$ and $v^h$. $W^q$, $W^k$, and $W^v \in \mathbb{R}^{dim \times dim}$ are the projection matrices. This means that the original image token is used as the query in the attention calculation, while the multi-frequency token serves as the key and value. Since our cross attention only applies to tokens in the same position of the sequence, we only have a small attention score matrix, which means a small computational cost.

Finally, we stack $\tilde{v}$ to obtain the final visual token sequence $\tilde{V}$, which is aligned with the text tokens through the connector. These tokens are then concatenated and fed into the LLM backbone to generate the response. 
\subsection{Training \& Inference}
\paragraph{Training}
Our training method follows the setting used in LLaVA~\cite{liu2024visual}, consisting of two stages: the pre-training (PT) stage and the supervised-finetuning (SFT) stage. During the PT stage, only the connector and the three projection matrices ($W_q,W_k,W_v$) are trained. In the SFT stage, the model connector, the three projection matrices in \cref{eq:train_main}, and the LLM backbone are trained. Compared to LLaVA, we introduce very few additional training parameters (only $W_q,W_k,W_v$, $\approx$ 3M).
\paragraph{Inference}
During model inference, we introduce an attenuation factor $\gamma$ in the multi-frequency feature extraction to modulate the strength of low- and high-frequency features. This adjustment helps suppress redundant frequency-domain features, thereby reducing hallucinations. Specifically, we rewrite the process in the \cref{eq:get_freq} as follows:
\begin{equation}
    \begin{cases}
  \mathcal{F}^l_c(u,v) = \mathcal{F}_c(u,v) \cdot \mathcal{H}^l_c(u,v) \cdot \mathrm G(\gamma)\\
  \mathcal{F}^h_c(u,v) = \mathcal{F}_c(u,v) \cdot \mathcal{H}^h_c(u,v) \cdot \mathrm G(\gamma).
    \end{cases}
\end{equation}
Where $G(\gamma)$ is a matrix whose values are obtained by randomly sampling from the uniform distribution $U(0,\gamma)$, where $\gamma\leq1$.

%% file: sec/4_Experiment.tex
\section{Experiment}
\subsection{Setup}
\paragraph{Implementation Detail} Similar to most work in the field of MLLM hallucination, we apply our proposed method on LLaVA-1.5-7B~\cite{liu2024improved} for our experiments. The LCS-558k dataset~\cite{liu2024improved} is used during the pre-training phase, while the LLaVA-mixed-665k dataset~\cite{liu2024improved} is used during the SFT stage. As~\cref{training_detail}, We keep the training parameters consistent with LLaVA-1.5. We set $\mathcal{D}_0$=30. At inference time, we employ sampling decoding with temperature=0.2. The codebase framework is PyTorch~\cite{paszke2019pytorch}, and experiments are conducted with 8$\times$H100 GPUs.
\paragraph{Baseline}
We select several well-established methods as baselines. DoLa~\cite{chuang2023dola} is a simple decoding strategy that mitigates hallucinations in LLMs by contrasting logits from different layers, without relying on external knowledge or fine-tuning. ITI~\cite{li2023inference} improves LLM truthfulness by modulating activations along specific attention head directions during inference. VCD~\cite{leng2024mitigating} is a training-free approach that contrasts output distributions from original and perturbed visual inputs to minimize bias and unimodal priors. ICD~\cite{wang2024mitigatinghallucinationslargevisionlanguage}contrasts distributions from standard and instruction disturbance, thereby increasing alignment uncertainty and effectively subtracting hallucinated concepts from the original distribution. SID~\cite{huo2024self} reduces vision-and-text association hallucinations by selectively preserving only the least important vision tokens early in decoding. AGLA~\cite{an2024agla} is a training-free method that mitigates object hallucinations by leveraging global features for response generation and local features for visual discrimination. OPERA~\cite{huang2024opera} introduces a novel MLLM decoding strategy that reduces hallucinations through an over-trust penalty and a retrospection-allocation mechanism. DOPRA~\cite{wei2024dopra} addresses hallucinations by applying weighted layer penalties and redistribution during decoding. HALC~\cite{chen2024halc} enhances vision-language tasks by incorporating fine-grained visual information and integrating both local and global contexts. CCA-LLaVA~\cite{xing2025mitigating} proposes Concentric Causal Attention (CCA), a positional alignment strategy that mitigates RoPE’s long-term decay by reducing the relative distance between visual and instruction tokens in MLLMs.
\subsection{Benchmark and Metric}
\paragraph{CHAIR~\cite{rohrbach2018object}}
Caption Hallucination Assessment with Image Relevance (CHAIR) is a widely used benchmark for assessing hallucinations in image captioning tasks. It works by creating a set of ground-truth object labels for each image, where any object mentioned in the caption that is not present in the label set is considered a hallucinated object. CHAIR consists of two variants: CHAIR$_i$ (instance-level) and CHAIR$_s$ (sentence-level), which are calculated as follows:
\begin{equation}
\mathrm{CHAIR_i} = \frac{|\{\mathrm{hallucinated 
\ objects}\}|}{|\{\mathrm{all \ mentioned 
\ objects}\}|},
\end{equation}
\begin{equation}
\mathrm{CHAIR_s} = \frac{|\{\mathrm{captions \ with \ hallucinated \ object}\}|}{|\{\mathrm{all \ captions }\}|}.
\end{equation}
Consistent with the evaluation settings of previous work~\cite{leng2024mitigating}, we randomly sample 500 images from the MSCOCO 2014 validation set~\cite{lin2014microsoft}, set the max-tokens to 512, and use the prompt ``Please describe this image in detail.'' In addition to the CHAIR metric, we also report F1 scores to assess the completeness of the generated image descriptions.
\paragraph{POPE~\cite{li2023evaluating}}
The Polling-based Object Probing Evaluation (POPE) is a benchmark created to evaluate object hallucinations within the VQA framework. It works by asking MLLMs targeted questions like ``Is there a \{object\} in the image?'', where \{object\} refers to a ground-truth item selected from three distinct splits. In the ``random'' split, objects are randomly chosen from the entire dataset. The ``popular'' split includes the most frequently occurring objects, and the ``adversarial'' split presents objects that are closely related to those in the image. For our analysis, we use the COCO dataset~\cite{lin2014microsoft}, selecting 500 images and posing six questions per image for each POPE split. We use the average F1 score of the three split sets as the evaluation metric.
\paragraph{MME~\cite{fu2024mmecomprehensiveevaluationbenchmark}}
In line with prior methods~\cite{leng2024mitigating,chuang2023dola,huo2024self}, we employ the hallucination subset of MME to systematically assess the model’s performance. This subset enables a detailed evaluation of both object-level and attribute-level hallucinations. For object-level hallucination, we employ the existence and count subsets, which enable the evaluation of a model’s ability to correctly identify the presence and quantity of objects within an image. Meanwhile, attribute-level hallucinations are examined using the position and color subsets, allowing us to assess how well MLLMs capture spatial relationships and color attributes.

\paragraph{MMBench~\cite{liu2024mmbench}}
To assess whether the overall capabilities of MLLMs are well preserved, we employ MMbench, a widely recognized benchmark specifically designed for evaluating the comprehensive abilities of MLLMs. MMbench serves as an objective and standardized evaluation framework, enabling a rigorous assessment of model performance across diverse multimodal tasks. It encompasses a meticulously curated set of over 3,000 multiple-choice questions, systematically categorized into 20 distinct ability dimensions, including but not limited to object localization, social reasoning, spatial understanding, and commonsense inference. By covering a broad spectrum of cognitive and perceptual skills, MMbench provides a holistic measure of an MLLM’s capacity, ensuring a robust and fair comparison across different models.
\subsection{Main Result}
\paragraph{Comparison of MFP with existing methods} 
\input{table/main_result}
To comprehensively evaluate the effectiveness of our proposed method, we compare its performance against various existing approaches across multiple evaluation metrics, as presented in \cref{main_result}. Our method consistently outperforms the baseline and other competing methods in key aspects.

First, in terms of POPE, our method achieves the highest F1 score of 86.2, surpassing the baseline (85.9) and all other methods, including DOPRA (85.6), SID (85.1), and CCA-LLAVA (85.5). Regarding CHAIR$_s$, our method attains a significantly lower score of 41.2, indicating mitigated sentence-level 
 object hallucinations compared to the baseline (50.2) and most other methods, such as VCD (51.0) and DoLa (57.0). Notably, our method also outperforms AGLA (43.0) and CCA-LLAVA (43.0). For CHAIR$_i$, our method achieves a score of 11.7, which is competitive with the best-performing method (CCA-LLAVA, 11.5). This suggests that our method effectively mitigates instance-level  hallucinations. Furthermore, the image captions generated by our model also maintain completeness relative to the baseline, as evidenced by the CHAIR F1 score (77.6 vs. 76.8). On MME, our method achieves the highest scores in Existence (195.0), Count (150.0), and Position (138.3), demonstrating strong performance in reducing hallucinations related to object presence, quantity, and spatial arrangement. With an overall MME score of 643.3, our approach ranks first, outperforming strong baselines such as VCD (604.6) and OPERA (592.3), further validating its effectiveness in hallucination mitigation. We provide more hallucination evaluation results in~\cref{appb} and~\cref{appc}.

 Second, in terms of MMBench, our method achieves the highest accuracy of 68.2, outperforming all other approaches, including SID (65.1), OPERA (64.4), and DoLa (63.8). In addition, we present the comparison results of our method with baselines on nine benchmarks evaluating generic capabilities, as detailed in~\cref{appa}. This demonstrates that our method not only preserves but also enhances the capabilities of the baseline model. 

In summary, our method demonstrates competitive performance  in mitigating object hallucinations while maintaining and improving the baseline model’s general capabilities.

\paragraph{Generalization of MFP across different architectures}
\input{table/general_result}
\Cref{general_result} presents the evaluation of the proposed MFP method across various model configurations, including different LLMs, visual encoders, input resolutions, and model sizes. The results consistently show that applying MFP improves performance within each architecture, as indicated by increased POPE scores and reduced CHAIR$_s$ and CHAIR$_i$ values. For LLMs , MFP leads to a slight increase in POPE (\eg, 85.9 to 86.2 in vicuna1.5-7B~\cite{zheng2023judgingllmasajudgemtbenchchatbot}, 85.4 to 86.2 in llama2-7B~\cite{touvron2023llama}) while significantly reducing CHAIR$_s$ and CHAIR$_i$, confirming its effectiveness across different LLM backbones. Similarly, for model sizes, MFP provides consistent improvements for both 7B and 13B models. For example, in vicuna1.5-13B, MFP reduces CHAIR$_s$ from 42.3 to 37.6, demonstrating its ability to mitigate hallucinations even in larger models. For visual encoders, MFP remains effective when applied to both CLIP~\cite{radford2021learningtransferablevisualmodels} and SigLIP~\cite{zhai2023sigmoid}. Within each encoder, MFP maintains a high POPE score while reducing CHAIR$_s$ and CHAIR$_i$. Furthermore, MFP demonstrates robustness across different input resolutions. Regardless of whether the input resolution is 336, 384, or 672, applying MFP consistently improves POPE (e.g., from $86.4$ to $86.8$ in LLaVA-Next~\cite{liu2024llavanext} when increasing resolution from 336 to 672) while maintaining competitive reductions in CHAIR$_s$ and CHAIR$_i$. These results confirm that MFP is architecture-agnostic, providing consistent improvements across different LLMs, visual encoders, input resolutions, and model sizes. This reinforces MFP as a reliable method for mitigating hallucinations in multimodal models, regardless of the architecture.

\paragraph{Compatibility of MFP with existing SOTA method}
As our method mitigates hallucinations from a novel perspective, it can be seamlessly integrated with existing method in an orthogonal manner. To our knowledge, PAI~\cite{liu2024paying} is the current SOTA method without additional data, achieving the best performance on the CHAIR benchmark. PAI amplifies image token attention and adjusts logits to reduce text bias, helping MLLMs focus more on visual information and mitigate hallucinations. However, because PAI operates at both the decoding and attention layers, it cannot be combined with many other hallucination mitigation methods (\eg, VCD, OPERA). Nonetheless, as a training-free inference-time approach, PAI can be effectively integrated with our proposed training-time method.
The experimental results are presented in \cref{sota_result}. Both PAI and MFP individually achieve strong performance. When using MFP alone, $\text{CHIAR}_s$ is reduced by 9.0 and $\text{CHIAR}_i$ by 3.3. Similarly, PAI alone decreases $\text{CHIAR}_s$ by 25.6 and $\text{CHIAR}_i$ by 7.2. However, combining PAI with MFP yields even better results, further reducing $\text{CHIAR}_s$ by 6.6 and $\text{CHIAR}_i$ by 2.0 compared to PAI alone, while maintaining the F1 score at 74.4. This combined method establishes a new state-of-the-art performance on the CHIAR benchmark. These results demonstrate the strong compatibility of our method.

\input{table/sota_result}
\subsection{Sensitivity Analysis}
\begin{figure}[t]
  \centering
   \includegraphics[width=0.95\linewidth]{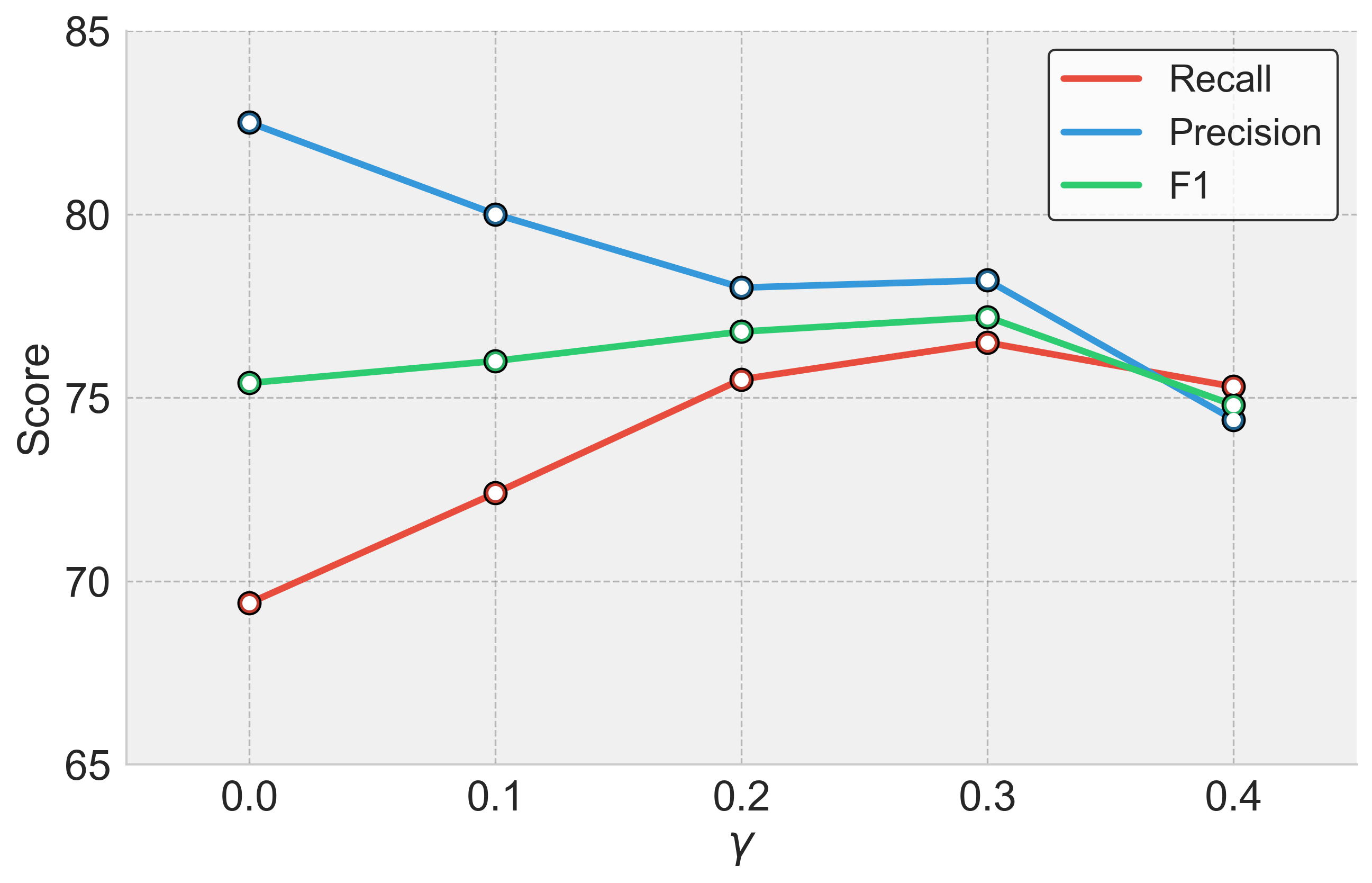}
   \caption{Results of sensitivity analysis on CHAIR benchmark for the parameter $\gamma$. The experiments are conducted on the LLaVA-1.5-7B model.
}
   \label{fig:sensitive}
\end{figure}
We choose the parameter $\gamma$ at inference time for sensitivity analysis, using recall, precision, and F1 score from the CHAIR benchmark as evaluation metrics. A lower $\gamma$ means fewer low- and high-frequency features are injected, and vice versa. As shown in~\cref{fig:sensitive}, when $\gamma$ decreases from 0.4 to 0.3, recall, precision, and F1 score all increase, reaching their peak values of 76.5, 78.2, and 77.2, respectively. This improvement suggests that our method effectively suppresses redundant high- and low-frequency features, thereby enhancing the model’s object detection capacity. However, when $\gamma$ is further reduced from 0.3 to 0.0, recall exhibits a decreasing trend (from 76.5 to 69.4), while precision increases (from 78.2 to 82.5). The F1 score initially improves slightly at $\gamma = 0.2$ before declining from 77.2 to 75.4 at $\gamma = 0.0$. This indicates that while reducing multi-frequency features alleviates object hallucination and improves precision, it may also lead to the loss of essential frequency-domain information, negatively impacting object detection and reducing recall. The observed trade-off between recall and precision highlights the importance of selecting an optimal $\gamma$ to balance feature suppression and retention. This result supports the conclusion that the core mechanism of our proposed method is to reduce redundant frequency-domain features, thereby improving overall model robustness.

\subsection{Ablation Study}
\input{table/ablation_result}
To further investigate the effectiveness of our proposed Multi-Frequency Processing (MFP) method, we conduct an ablation study by removing different frequency features and the inference-time parameter $\gamma$, as shown in \cref{tab:ablation_result}. Compared to the baseline, MFP significantly reduces CHAIR$_s$ and CHAIR$_i$ scores from 50.2 to 41.2 and from 15.0 to 11.7, respectively, demonstrating its effectiveness in mitigating object hallucinations. When low-frequency features are removed during training (w/o low), CHAIR$_s$ and CHAIR$_i$ increase to 49.8 and 14.8, respectively, suggesting that low-frequency features play a crucial role in suppressing spurious detections. Similarly, removing high-frequency features (w/o high) results in a CHAIR$_s$ score of 51.4 and a CHAIR$_i$ score of 14.4, indicating that high-frequency features also contribute to hallucination suppression, though their impact is slightly less pronounced. Notably, both settings perform worse than the Multi-Frequency features, highlighting that integrating both low- and high-frequency features is more effective than relying on either features alone. Furthermore, when $\gamma$ is removed at inference time (w/o $\gamma$), CHAIR$_s$ and CHAIR$_i$ degrade to 51.2 and 14.3, respectively, underscoring the importance of decaying frequency features during inference. Importantly, across all ablation settings, the POPE score remains consistently higher than the baseline, ranging from 86.2 to 86.7, indicating that our method robustly increases POPE score regardless of specific frequency feature being removed. These results validate that while both low- and high-frequency features contribute to hallucination mitigating, their joint utilization within MFP leads to the most effective mitigating.

%% file: table/main_result.tex
\begin{table*}
  \centering
  \setlength\tabcolsep{4pt}
  \begin{tabular}{lccccccccccc}
    \toprule
    \multirow{2}{*}{Method} & \multicolumn{1}{c}{POPE}  & \multicolumn{4}{c}{CHAIR } &\multicolumn{5}{c}{MME} & \multicolumn{1}{c}{MMB} \\
     \cmidrule(lr){2-2} \cmidrule(lr){3-6} \cmidrule(lr){7-11} \cmidrule(lr){12-12}
    & F1  & {CHAIR\( _{s}^\downarrow\)} &{CHAIR\( _{i}^\downarrow \)} & F1 & {Avg. Len} &Existence &Count &Position & Color &Overall &Acc.\\
    \cmidrule{1-1}\cmidrule(lr){2-2} \cmidrule(lr){3-6} \cmidrule(lr){7-11} \cmidrule(lr){12-12}
    Baseline        & 85.9 & 50.2   & 15.0  & 76.8 & 99.2 &175.7 &124.7 &114.0 &51.0 &565.3 &63.0 \\
    DoLa~\cite{chuang2023dola}          & 80.2 & 57.0   & 15.2  & - & 97.5  &175.0 &108.3 &90.0 &138.3 &511.7 &63.8 \\
    ITI~\cite{li2023inference}    & 83.7 & 48.2   & 13.9  & - & 98.6  &- &- &- &- &- &- \\
    VCD~\cite{leng2024mitigating}            & 83.2 & 51.0   & 14.9  & - & 101.9  &184.7 &138.3 &\underline{128.7} &53.0 &604.7 &63.9 \\    ICD~\cite{wang2024mitigatinghallucinationslargevisionlanguage}           &- & 47.4   & 13.9  & - & -  &185.0 &117.9 &117.5 &162.1 &582.5  &63.1 \\

    SID~\cite{huo2024self}           & 85.1 & 45.0   & 11.7  & - & -   &190.0 &148.3 &128.3 &\underline{175.0} &641.7 &65.1 \\ 
    AGLA~\cite{an2024agla}  &   84.6 & \underline{43.0}   & 14.1 & 78.9 & 98.8  &- &- &- &- &- &- \\
    OPERA~\cite{huang2024opera}         & 85.2 & 47.0   & 14.6  & - & 95.3  &180.7 &133.3 &123.3 &155.0 &592.3 &64.4 \\
    DOPRA~\cite{wei2024dopra} & \underline{85.6} & 46.3   & 13.8  & - & 96.1  &- &- &- &- &- &- \\
    HALC~\cite{chen2024halc}  & 83.9 & 50.2 & 12.4 & - & 97.2 &- &- &- &- & &- \\
    CCA-LLaVA~\cite{xing2025mitigating}  & 85.5 & 43.0   & \textbf{11.5}  & - & 96.6  &\underline{190.0} &\underline{148.3} &128.3 &\textbf{175.0} &\underline{641.7} &\underline{65.4} \\
    \textbf{Ours} &\textbf{86.2} &\textbf{41.2} &\underline{11.7} &77.6 &94.4  &\textbf{195.0} &\textbf{150.0} &\textbf{138.3} &160.0 &\textbf{643.3} &\textbf{68.2} \\

    \bottomrule
  \end{tabular}
    \caption{Compare results of MFP with other SOTA methods on POPE and CHAIR datasets. We set $\gamma=0.23$. The evaluation results of the compared methods are from published papers. The best performances within each metric are \textbf{bolded}. The second best performances are \underline{underlined}.}
      \label{main_result}
\end{table*}

%% file: table/general_result.tex
\begin{table}
  \centering
  \small
  \setlength\tabcolsep{2pt}
  \begin{tabular}{lclcccc}
    \toprule
    LLaVA & V. E.  &LLM &Res. &POPE &CHAIR$_{s}^\downarrow$ &CHAIR$_{i}^\downarrow$\\
    \midrule
    V1.5& CLIP & vicuna1.5$^{\text{7B}}$ &336 &85.9 &50.2 &15.0 \\
    w/ MFP & CLIP & vicuna1.5$^{\text{7B}}$ &336 &\textbf{86.2} &\textbf{41.2} &\textbf{11.7} \\
    \midrule
    V1.5& CLIP & vicuna1.5$^{\text{13B}}$ &336 &85.9 &53.0 &14.6 \\
    w/ MFP & CLIP & vicuna1.5$^{\text{13B}}$ &336 &\textbf{86.4} &\textbf{37.6} &\textbf{10.4} \\
    \midrule
    V1.5& SigLIP & vicuna1.5$^{\text{7B}}$ &384 &\textbf{86.4} &47.0 &12.0 \\
    w/ MFP & SigLIP & vicuna1.5$^{\text{7B}}$ &384 &85.4 &\textbf{42.0} &\textbf{11.9} \\
    \midrule
    V1.5& CLIP & llama2$^{\text{7B}}$ & 336 &85.4 &48.0 &14.8 \\
    w/ MFP & CLIP & llama2$^{\text{7B}}$ & 336 &\textbf{86.2} &\textbf{46.0} &\textbf{14.1} \\
    \midrule
    Next& CLIP &vicuna1.5$^{\text{7B}}$ &672 &86.4 &51.0 &12.8 \\
    w/ MFP & CLIP &vicuna1.5$^{\text{7B}}$ &672 &\textbf{86.8} &\textbf{45.4} &\textbf{11.7} \\
    \bottomrule
  \end{tabular}
    \caption{Results of MFP cross different architectures. The best performances within each setting are \textbf{bolded}. V. E. refer to visual encoder and Res. refer to resolution. From top to bottom we set $\gamma$ to 0.23, 0.1, 0.1, 0.3 and 0.4. Considering that the data of LLaVA-Next is not publicly available, we train LLaVA-Next using the data of LLaVA-1.5.}
      \label{general_result}
      
\end{table}

%% file: table/sota_result.tex
\begin{table}
  \centering
  \small
  \setlength\tabcolsep{4pt}
  \begin{tabular}{cccccc}
    \toprule
    \multirow{1}{*}{\bf Training} & \multicolumn{1}{c}{\bf Inference}        & {CHAIR\( _{s}^\downarrow\)} &{CHAIR\( _{i} ^\downarrow\)} & F1 & {Avg. Len}\\
    \midrule
    Vanilla & Vanilla   & 50.2   & 15.0  & 76.8 & 99.2  \\
    \midrule
    MFP &Vanilla &41.2 &11.7 &77.6 &94.4 \\
    Vanilla &PAI &24.6 &7.2 &74.4 &87.6 \\
    MFP & PAI&\textbf{18.0} &\textbf{5.2} &74.4 &68.8 \\

    \bottomrule
  \end{tabular}
    \caption{Results of compatibility of MFP with existing SOTA method on CHAIR benchmark. $\gamma=0.23$. The best performances within each setting are \textbf{bolded}. The experiments are conducted on the LLaVA-1.5-7B model.}
      \label{sota_result}
\end{table}

%% file: table/ablation_result.tex
\begin{table}[t]
\centering
  \setlength\tabcolsep{5pt}

\begin{tabular}{lcccc}
    \toprule
    Method  & Stage      &POPE & {CHAIR\( _{s}^\downarrow\)} &{CHAIR\( _{i}^\downarrow \)} \\
    \midrule
      Baseline & - & 85.9 & 50.2  & 15.0  \\
    \midrule
     MFP   & -    & 86.2 & \textbf{41.2}  & \textbf{11.7}  \\
    
    w/o low & Training           & \textbf{86.7}& 49.8  & 14.8  \\
    w/o high  & Training         & 86.5 & 51.4   & 14.4\\

    w/o $\gamma$ & Inference     & 86.5 & 51.2  & 14.3  \\ \hline
\end{tabular}
    \caption{Result of ablating different parts of MFP. The best performance is \textbf{bolded}. $\gamma$=0.23 is set for all experiments. The experiments are conducted on the LLaVA1.5-7B model.
}
      \label{tab:ablation_result}
\end{table}

%% file: sec/5_Conclusion.tex
\section{Conclusion}
In this paper, we conduct the first analysis of object hallucinations from the perspective of frequency-domain in MLLMs, revealing that these models tend to be over-susceptible to specific frequency features when recognizing objects. To mitigate this issue, we propose MFP, a simple, cost-effective, and easily pluggable method that effectively mitigates object hallucinations. Our method demonstrates competitive performance across multiple hallucination benchmarks and generalizes well across different model architectures.
Furthermore, as a training-time method, MFP can be seamlessly integrated with inference-time method, even achieving state-of-the-art performance on the CHAIR benchmark. These findings underscore the potential of frequency-domain techniques for hallucination mitigation and highlight the importance of further investigating the relationship between image features and object recognition in MLLMs. We hope that our work inspires future research to develop more powerful frequency-based methods to enhance the reliability of MLLMs and mitigate object hallucinations.

%% file: sec/appendix.tex
\newpage
\section*{Appendix}
\section{Comparison on More General Benchmarks}
\label{appa}
\Cref{tab:general} illustrates the performance comparison between our proposed MFP method and the baseline model across 9 general benchmarks. Our evaluation is conducted on a diverse set of benchmarks, including TextVQA~\cite{singh2019vqamodelsread}, VizWiz~\cite{gurari2018vizwizgrandchallengeanswering}, MMBench (English \& Chinese), MME, MM-Vet~\cite{yu2024mmvetevaluatinglargemultimodal}, ScienceQA~\cite{lu2022learnexplainmultimodalreasoning}, SEED-Image~\cite{li2023seedbenchbenchmarkingmultimodalllms}, and VQAv2~\cite{goyal2017makingvvqamatter}, covering various aspects of multimodal understanding and reasoning. 

Overall, MFP achieves a comparable performance to the baseline, with an average score of 61.2, slightly surpassing the baseline's 61.0. While MFP demonstrates improvements on certain benchmarks, such as MMBench (EN \& CN) and ScienceQA, it exhibits minor performance drops on others, including VQAv2 and MME. According to the evaluation results, our model maintains the same general capability as the baseline.
\input{table/general_benchmarks}

\section{GPT-4o Assistant Evaluation}
In keeping with PAI, we also chose to use GPT-assisted evaluation. Since GPT-4V has been removed from the market, we chose the more advanced GPT-4o to evaluate our model. We construct prompts and input both the images and the description responses from two assistants into GPT-4o as follow:
\begin{tcolorbox}[sharp corners, boxrule=0.5pt, breakable]
\{image\}\\You are required to score the performance of two AI assistants in describing a given
image. You should pay extra attention to the hallucination, which refers to the part of
descriptions that are inconsistent with the image content, such as claiming the existence
of something not present in the image or describing incorrectly in terms of the counts,
positions, or colors of objects in the image. Please rate the responses of the assistants
on a scale of 1 to 10, where a higher score indicates better performance, according to
the following criteria:

1: Accuracy: whether the response is accurate with respect to the image content. Responses with fewer hallucinations should be given higher scores.

2: Detailedness: whether the response is rich in necessary details. Note that hallucinated
descriptions should not count as necessary details.

Please output the scores for each criterion, containing only two values indicating the
scores for Assistant 1 and 2, respectively. The two scores are separated by a space.
Following the scores, please provide an explanation of your evaluation, avoiding any
potential bias and ensuring that the order in which the responses were presented does
not affect your judgment.\\

[Assistant 1] 

\{Response of Assistant 1\}

[Assistant 2] 

\{Response of Assistant 2\}\\

Output format:\\

Accuracy: \{Scores\}

Reason:

Detailedness: \{Scores\}

Reason:
\end{tcolorbox}
The evaluation considers two key dimensions: Accuracy and Detailedness Our evaluation Settings are consistent with PAI, and the results are shown in \cref{tab:GPT}. Compared with the baseline and VCD, our method has better accuracy and detailedness.
\label{appb}
\input{table/vcdicd}

\section{More Training Details}
\input{table/training_setup}
\label{training_detail}
For main results, we keep all training hyperparameters roughly the same as the LLaVA series. \Cref{tab:training_setup} presents a detailed training setup during PT and SFT stage.

\section{Case Study}
\label{appc}
\begin{figure*}
    \centering
    \begin{subfigure}{0.80\linewidth}
        \centering
        \includegraphics[width=\linewidth]{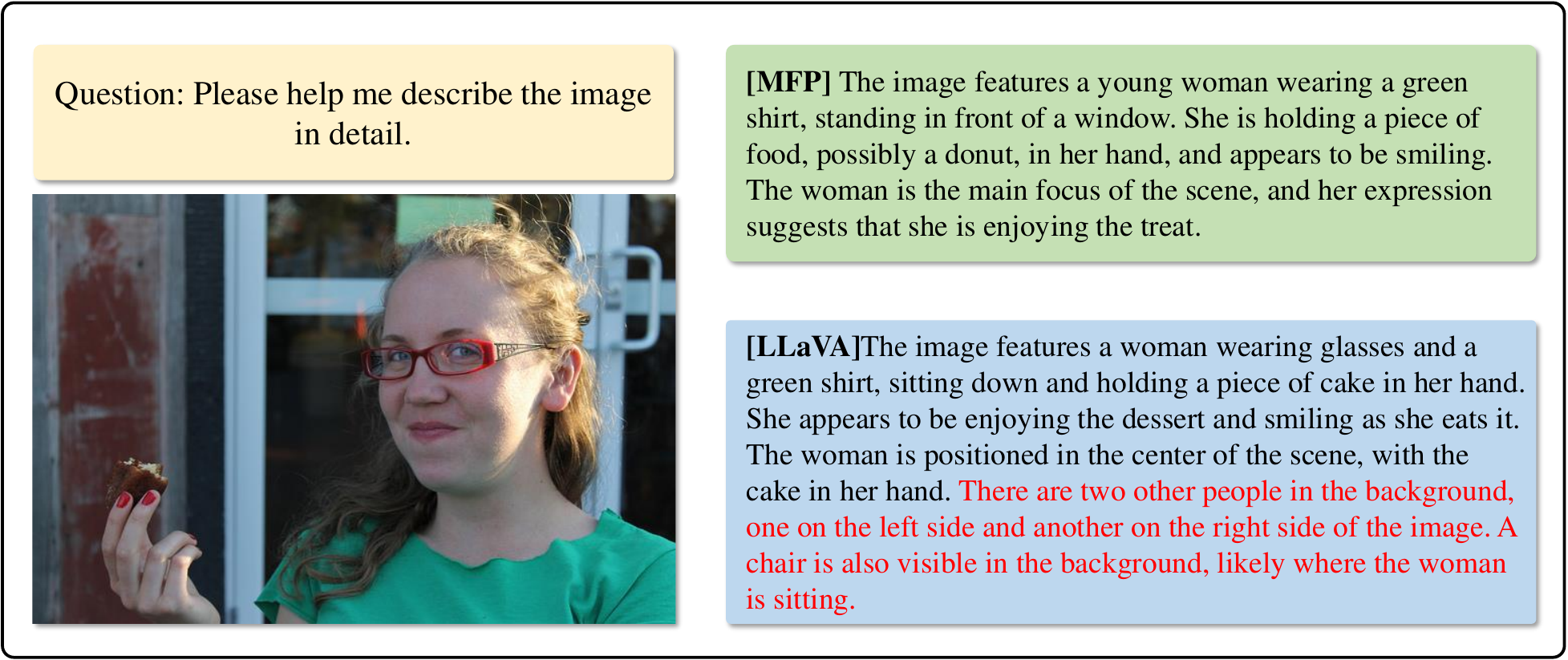}
    \end{subfigure}
    \begin{subfigure}{0.80\linewidth}
        \centering
        \includegraphics[width=\linewidth]{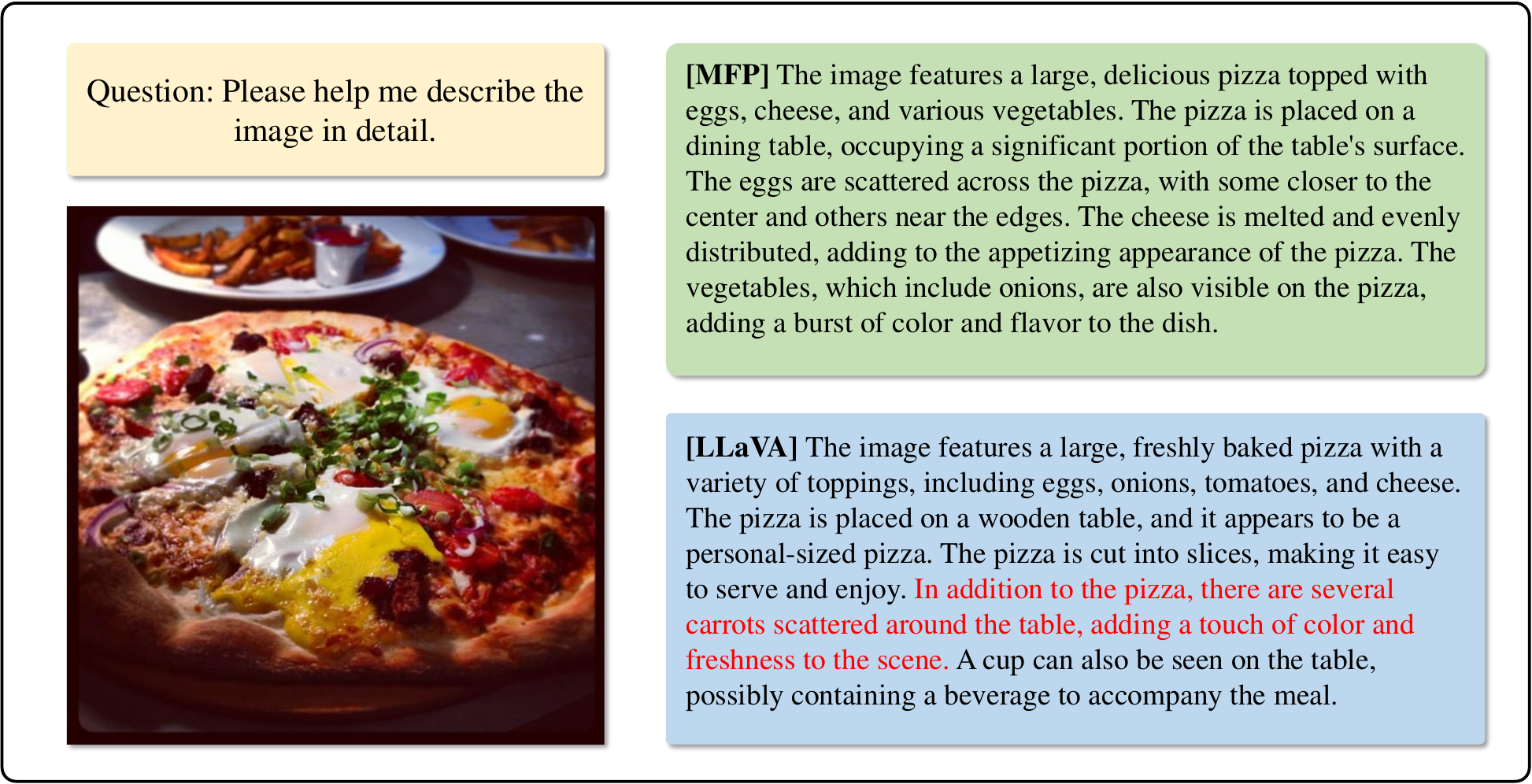}
    \end{subfigure}
    \begin{subfigure}{0.80\linewidth}
        \centering
        \includegraphics[width=\linewidth]{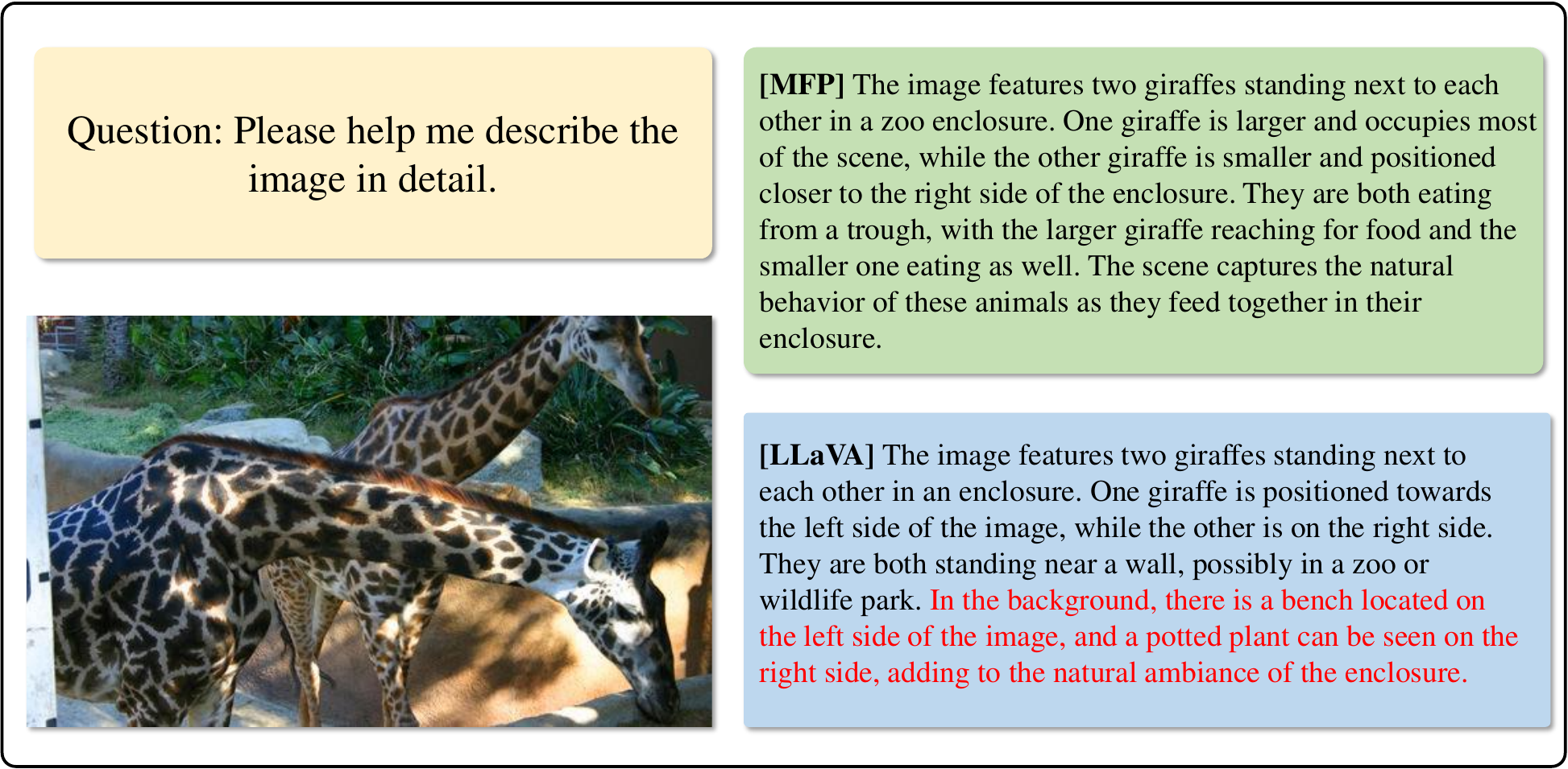}
    \end{subfigure}
    \caption{Comparison between our proposed MFP method and the original output in some cases. The hallucinating responses are highlighted in red.}
    \label{fig:case_study}
\end{figure*}

\Cref{fig:case_study} presents a case-by-case comparison between our proposed MFP method and the original output. The results demonstrate that our approach significantly reduces hallucinations. 

\section{Limitation}
Due to time and computational constraints, our evaluation is conducted on a limited set of model architectures, which may impact the generalizability of our findings to a broader range of architectures. Additionally, our experiments do not include a comprehensive analysis of key hyperparameters, such as the number of feature fusion layers and the specific frequency ranges utilized. A more systematic exploration of these factors could provide deeper insights into their influence on performance and potentially lead to further improvements.

%% file: table/general_benchmarks.tex
\begin{table*}[ht]
\begin{tabular}{l|ccccccccc|c}
\toprule
Model    & VQA$^{\rm{T}}$ & VizWiz & MMB$^{\rm{EN}}$ &MMB$^{\rm{CN}}$& MME & MM-Vet & SQA$^{\rm{I}}$ & SEED$^{\rm{I}}$ & VQA$^{\rm{v2}}$ & Overall \\ 
\midrule
Baseline  & \textbf{58.6}   &   50.0      &   64.3     &    58.3         &   \textbf{75.5}  &    \textbf{30.5}    &     66.8      &\textbf{66.1}     &  \textbf{78.5}     &   61.0      \\ 
MFP &   56.5      &   \textbf{53.4}     &     \textbf{68.2}  &     \textbf{59.0}   &   73.3     &    30.4       &   \textbf{67.7}   &   65.1    &   76.9  & \textbf{61.2}   \\

\bottomrule
\end{tabular}
\caption{Result of General Benchmarks. The best performance is \textbf{bolded}.}
\label{tab:general}
\end{table*}

%% file: table/vcdicd.tex
\begin{table}
  \centering
  \scalebox{1.0}{
  \begin{tabular}{ccc}
    \toprule
    \textbf{Method}& Accuracy$\uparrow$&Detailedness$\uparrow$\\
    \midrule
    Baseline& 5.38& 5.88\\
    VCD&5.83&5.93\\
    MFP&\textbf{6.29}&\textbf{6.52}\\
    \bottomrule
  \end{tabular}}
  \caption{Result of GPT-4o Assistant Evalution. The best performance is \textbf{bolded}.}
\label{tab:GPT}
\end{table}

%% file: table/training_setup.tex
\begin{table}[ht]
\centering
\begin{tabular}{l| c c}
\toprule
Hyperparameter & PT & SFT \\
\midrule
batch size & 256 & 128 \\
lr & 1e-3 & 2e-5 \\
lr schedule & \multicolumn{2}{c}{cosine decay} \\
lr warmup ratio & \multicolumn{2}{c}{0.03} \\
weight decay & \multicolumn{2}{c}{0} \\
epoch & \multicolumn{2}{c}{1} \\
optimizer & \multicolumn{2}{c}{AdamW} \\
DeepSpeed stage & 2 & 3 \\
\bottomrule
\end{tabular}
\caption{
Hyperparameters of our model's pretrain and finetune.
}
\vspace{2mm}
\label{tab:training_setup}
\end{table}